%% file: main.tex
\documentclass{article}
\pdfoutput=1

 \usepackage[preprint]{neurips_2021}

\usepackage[utf8]{inputenc} 
\usepackage[T1]{fontenc}    
\usepackage{url}            
\usepackage{booktabs}       
\usepackage{amsfonts}       
\usepackage{nicefrac}       
\usepackage{microtype}      
\usepackage{xcolor}         
\usepackage{enumitem}

\definecolor{darker}{rgb}{0,0.15,0.7}
\usepackage[colorlinks, urlcolor=darker, citecolor=darker, linkcolor=darker]{hyperref}
\setcitestyle{round}

\input{math_commands}

\usepackage{amsmath}
\usepackage{amssymb}
\usepackage{graphicx}
\usepackage{multirow}
\usepackage{makecell}

\usepackage{caption}
\usepackage{subfig}
\usepackage{tikz}

\newcommand{\platent}{p_{\mathrm{latent}}}
\newcommand{\pdata}{p_{\mathrm{data}}}

\title{On Fast Sampling of Diffusion Probabilistic 
Models 
}

\author{%
    Zhifeng Kong \\
    Computer Science and Engineering \\
    UC San Diego \\
    \texttt{z4kong@eng.ucsd.edu}
    \And
    Wei Ping \\
    NVIDIA \\
    \texttt{wping@nvidia.com}
}

\begin{document}

\maketitle

\begin{abstract}
  In this work, we propose FastDPM, a unified framework for fast sampling in diffusion probabilistic models.
  FastDPM generalizes previous methods and gives rise to new algorithms with improved sample quality. 
  We systematically investigate the fast sampling methods under this framework across different domains, on different datasets, and with different amount of conditional information provided for generation.
  We find the performance of a particular method depends on data domains~(e.g., image or audio), the trade-off between sampling speed and sample quality, and the amount of conditional information.
  We further provide insights and recipes on the choice of methods for practitioners.
\end{abstract}

\section{Introduction}

Diffusion probabilistic models are a class of deep generative models that use Markov chains to gradually transform between a simple distribution (e.g., isotropic Gaussian) and the complex data distribution~\citep{sohl2015deep, ho2020denoising}. 
Most recently, these models have obtained the state-of-the-art results in several important domains, including image synthesis~\citep{ho2020denoising, song2020score, dhariwal2021diffusion}, audio synthesis~\citep{kong2020diffwave, chen2020wavegrad}, and 3-D point cloud generation~\citep{luo2021diffusion, zhou20213d}.
We will use ``diffusion models'' as shorthand to refer to this
family of models.

Diffusion models usually comprise: \emph{i)} a {parameter-free} $T$-step Markov chain named the \emph{diffusion process}, which gradually adds random noise into the data, and \emph{ii)} a {parameterized} $T$-step Markov chain called the \emph{reverse} or \emph{denoising process}, which removes the added noise as a denoising function.
The likelihood in diffusion models is intractable, but they can be efficiently trained by optimizing a variant of the variational lower bound.
In particular, \citet{ho2020denoising} propose a certain parameterization called the denoising diffusion probabilistic model~(DDPM) and show its connection with denoising score matching~\citep{song2019generative}, so the reverse process can be viewed as sampling from a score-based model using Langevin dynamics.
DDPM can produce high-fidelity samples reliably with large model capacity and outperforms the state-of-the-art models in image and audio domains~\citep{dhariwal2021diffusion, kong2020diffwave}.
However, a noticeable limitation of diffusion models is their expensive denoising or sampling process. 
For example, DDPM requires a Markov chain with $T=1000$ steps to generate high quality image samples~\citep{ho2020denoising}, and DiffWave requires $T=200$ to obtain high-fidelity audio synthesis~\citep{kong2020diffwave}.
In other words, one has to run the forward-pass of the neural network $T$ times to generate a sample, which is much slower than the state-of-the-art GANs or flow-based models for image and audio synthesis~\citep[e.g.,][]{karras2020analyzing, kingma2018glow, kong2020hifi, ping2019waveflow}.

To deal with this limitation, several methods have been proposed to reduce the length of the reverse process to $S\ll T$ steps. 
One class of methods compute continuous noise levels based on discrete diffusion steps and retrain a new model conditioned on these continuous noise levels~\citep{song2019generative, chen2020wavegrad, okamoto2021noise, san2021noise}. Then, a shorter reverse process can be obtained by carefully choosing a small set~(size $S$) of noise levels. 
However, these methods cannot reuse the pretrained diffusion models, because the state-of-the-art DDPM models are conditioned on discrete diffusion steps~\citep[][]{ho2020denoising, dhariwal2021diffusion}. 
It is also unclear the diffusion models conditioned on continuous noise levels can achieve comparable sample quality as the state-of-the-art DDPMs on challenging unconditional image and audio synthesis tasks~\citep{dhariwal2021diffusion, kong2020diffwave}. 
Another class of methods directly approximate the original reverse process of DDPM models with shorter ones (of length $S$), which are conditioned on discrete diffusion steps~\citep{song2020denoising, kong2020diffwave}. 
Although both classes of methods have shown the trade-off between sampling speed and sample quality~(i.e., larger $S$ lead to higher sample quality), the fast sampling methods without retraining are more advantageous for fast iteration and deployment, while still keeping high-fidelity synthesis with small number of steps in the reverse process~(e.g., $S=6$ in \citet{kong2020diffwave}).

In this work, we propose FastDPM, a unified framework of fast sampling methods for diffusion models without retraining.
The core idea of FastDPM is to \emph{i)} generalize discrete diffusion steps to continuous diffusion steps, and \emph{ii)} design a bijective mapping between continuous diffusion steps and continuous noise levels. Then, we use this bijection to construct an approximate diffusion process and an approximate reverse process, both of which have length $S\ll T$. 

FastDPM includes and generalizes the fast sampling algorithms from denoising diffusion implicit models~(DDIM)~\citep{song2020denoising} and DiffWave~\citep{kong2020diffwave}. In detail, FastDPM offers two ways to construct the approximate diffusion process: selecting $S$ steps in the original diffusion process, or more flexibly, choosing $S$ variances. FastDPM also offers ways to construct the approximate reverse process: using the stochastic DDPM reverse process~(DDPM-rev), or using the implicit (deterministic) DDIM reverse process~(DDIM-rev). We can control the amount of stochasticity in the reverse process of FastDPM as in \citet{song2020denoising}.

FastDPM gives rise to new algorithms with improved sample quality than previous methods when the length of the approximate reverse process $S$ is small. We then extensively evaluate the family of FastDPM methods across image and audio domains. We find the deterministic DDIM-rev significantly outperforms the stochastic DDPM-rev in image generation tasks, but DDPM-rev significantly outperforms DDIM-rev in audio synthesis tasks. Finally, we investigate the performance of different methods by varying the amount of conditional information. We find with different amount of conditional information, we need different amount of stochasticity in the reverse process of FastDPM.

In summary, we make the following contributions:
\vspace{-0.4em}
\begin{enumerate}[leftmargin=3.1em]
    \item  
    FastDPM introduces the concept of continuous diffusion steps, and generalizes prior fast sampling algorithms without retraining \citep{song2020denoising,kong2020diffwave}.
    \vspace{-0.1em}
    \item 
    FastDPM gives rise to new algorithms with improved sample quality when the length of the approximate reverse process $S$ is small.
    \vspace{-0.1em}
    \item 
    We extensively evaluate FastDPM across image and audio domains, and provide insights and recipes on the choice of methods for practitioners.
    \vspace{-0.1em}
\vspace{-0.4em}
\end{enumerate}
We organize the rest of the paper as follows.
Section~\ref{sec:related_work} discusses related work.
We introduce the preliminaries of diffusion models in Section~\ref{sec:preliminaries}, and propose FastDPM in Section~\ref{sec:fast-dpm}.
We report experimental results in Section~\ref{sec:experiments} and conclude the paper in Section~\ref{sec:conclusion}.

\section{Related Work}
\label{sec:related_work}
Diffusion models are a class of powerful deep generative models~\citep{sohl2015deep, ho2020denoising, goyal2017variational}, which have received a lot of attention recently.
These models have been applied to various domains, including image generation~\citep{ho2020denoising, dhariwal2021diffusion}, audio synthesis~\citep{kong2020diffwave,chen2020wavegrad, okamoto2021noise}, image or audio super-resolution~\citep{li2021srdiff, lee2021nu}, text-to-speech~\citep{jeong2021diff, popov2021grad}, music synthesis~\citep{liu2021diffsinger, mittal2021symbolic},
3-D point cloud generation~\citep{luo2021diffusion, zhou20213d}, and language models~\citep{hoogeboom2021argmax}.
Diffusion models are connected with scored-based models~\citep{song2019generative, song2020improved, song2020score}, and there have been a series of research extending and improving diffusion models~\citep{song2020score,gao2020learning,dhariwal2021diffusion,san2021noise,meng2021improved}.

There are two families of methods aiming for accelerating diffusion models at synthesis, which reduce the length of the reverse process from $T$ to a much smaller $S$. 
One family of methods tackle this problem at training. They retrain the network conditioned on continuous noise levels instead of discrete diffusion steps~\citep{song2019generative, chen2020wavegrad, okamoto2021noise, san2021noise}. 
Assuming that the corresponding network is able to predict added noise at any noise level, we can carefully choose only $S \ll T$ noise levels and construct a short reverse process just based on them. 
\citet{san2021noise} present a learning scheme that
can step-by-step adjust those noise level parameters, for any given
number of steps $S$.
Another family of methods aim to directly approximate the original reverse process within the pretrained DDPM conditioned on discrete steps. In other words, no retraining is needed. \citet{song2020denoising} introduce denoising diffusion implicit models~(DDIM), which contain non-Markovian processes that lead to an equivalent training objective as DDPM. 
These non-Markovian processes naturally permit "jumping steps", or formally, using a subset of steps to form a short reverse process. However, compared to using continuous noise levels, selecting discrete steps offers less flexibility. \citet{kong2020diffwave} introduce a fast sampling algorithm by interpolating steps according to corresponding noise levels. This can be seen as an attempt to map continuous noise levels to discrete diffusion steps. 
However, it lacks both theoretical justification for the interpolation and extensive empirical studies.

In this paper, we propose FastDPM, a method that approximates the original DDPM model. FastDPM constructs a bijective mapping between  (continuous) diffusion steps and continuous noise levels. This allows us to take advantage of the flexibility of using these continuous noise levels. FastDPM generalizes \citet{kong2020diffwave} by using Gamma functions to compute noise levels, which naturally extends from discrete domain to continuous domain. FastDPM generalizes \citet{song2020denoising} by providing a special set of noise levels that exactly correspond to integer steps.

\section{Diffusion Models}
\label{sec:preliminaries}
Let $d$ be the data dimension. Let $\pdata$ be the data distribution and $\platent=\gN(0,I_{d\times d})$ be the latent distribution. Then, the denoising diffusion probabilistic model \citep[DDPM,][]{sohl2015deep, ho2020denoising} is a deep generative model consisting two Markov chains called diffusion and reverse processes, respectively. The length of each Markov chain is $T$, which is called the number of diffusion or reverse steps.  
The \emph{diffusion process} gradually adds Gaussian noise to the data distribution until the noisy data distribution is close to the latent distribution. Formally, the diffusion process from data $x_0\sim\pdata$ to the latent variable $x_T$ is defined as:
\begin{align}
q(x_1,\cdots,x_T|x_0) = \prod_{t=1}^T q(x_t|x_{t-1}),
\end{align}
where each of $q(x_t|x_{t-1})=\gN(x_t;\sqrt{1-\beta_t}x_{t-1},\beta_t I)$ for some small constant $\beta_t>0$. The hyperparameters $\beta_1,\cdots,\beta_T$ are called the \emph{variance schedule}.

The \emph{reverse process} aims to eliminate the noise added in each diffusion step. Formally, the reverse process from $x_T\sim\platent$ to $x_0$ is defined as:
\begin{align}\label{eq: reverse process}
p_{\theta}(x_0,\cdots,x_{T-1}|x_T)=\prod_{t=1}^T p_{\theta}(x_{t-1}|x_t),
\end{align}
where each of $p_{\theta}(x_{t-1}|x_t)$ is defined as $\gN(x_{t-1};\mu_{\theta}(x_t,t), \sigma_t^2I)$; the mean $\mu_{\theta}(x_t,t)$ is parameterized through a neural network and the variance $\sigma_t$ is time-step dependent constant. Based on the reverse process, the sampling process is to first draw $x_T \sim \gN(0,I)$, then draw $x_{t-1}\sim p_{\theta}(x_{t-1}|x_t)$ for $t=T,T-1,\cdots,1$, and finally outputs $x_0$.

The training objective of DDPM is based on the variational evidence lower bound (ELBO). Under a certain parameterization introduced by~\citet{ho2020denoising}, the objective can be largely simplified. 
One may first define constants
$\alpha_t = 1 - \beta_t$, $\bar{\alpha}_t = \prod_{i=1}^t\alpha_i$, $\tilde{\beta}_t=\frac{1-\bar{\alpha}_{t-1}}{1-\bar{\alpha}_t}\beta_t$ for $t>1$ and $\tilde{\beta}_1=\beta_1$.
Then, a noticeable property of diffusion model is that
\begin{align}
\label{q(x_t|x_0)}
q(x_t|x_0)=\gN(x_t;~ \sqrt{\bar{\alpha}_t}x_0,~ (1-\bar{\alpha}_t)  I), 
\end{align}
thus one can directly sample $x_t$ given $x_0$~(see derivation in Appendix~\ref{appendix:derivations}).

Furthermore, one may parameterize
$\mu_{\theta}(x_t, t) = \frac{1}{\sqrt{\alpha_t}}\left(x_t-\frac{\beta_t}{\sqrt{1-\bar{\alpha}_t}}\epsilon_{\theta}(x_t, t)\right)$,
where $\epsilon_{\theta}$ is a neural network taking $x_t$ and the diffusion-step $t$ as inputs. In addition, $\sigma_t$ is simply parameterized as $\tilde{\beta}_t^{\frac12}$ defined above.
\citet{ho2020denoising} show that minimizing the following unweighted variant of the ELBO leads to higher generation quality: 
\begin{equation}
\label{eq: train_obj}
\min_{\theta} L_\mathrm{{unweighted}}(\theta) = 
\mathbb{E}_{x_0,\epsilon,t}\ 
\|\epsilon - \epsilon_{\theta}(x_t, ~t)\|_2^2,
\end{equation}
where $\epsilon\sim\gN(0,I)$, $x_0\sim\qdata$, $t$ is uniformly taken from $1,\cdots,T$, and $x_t = \sqrt{\bar{\alpha}_t}\cdot x_0 + \sqrt{1-\bar{\alpha}_t}\cdot \epsilon$  from \eqref{q(x_t|x_0)}.
One may simply interpret this objective as a mean-squared error loss between the true noise $\epsilon$ and the predicted noise $\epsilon_{\theta}(x_t, ~t)$ at each time-step.

\section{FastDPM: A Unified Framework for Fast Sampling in Diffusion Models}
\label{sec:fast-dpm}
In order to achieve high-fidelity synthesis, the number of diffusion steps $T$ in DDPM is set to be very large so that $q(x_T|x_0)$ is close to $\platent$. For example, $T=1000$ in image synthesis \citep{ho2020denoising} and $T=200$ in audio synthesis \citep{kong2020diffwave}. Then, sampling from DDPM needs running through the network $\epsilon_{\theta}$ for as many as $T$ times, which can be very slow. 
In this section, we propose FastDPM, which approximates the pretrained DDPM via much shorter diffusion and reverse processes of length $S\ll T$, thus it can generate a sample by only running the network $S$ times.
The core idea of FastDPM is to: \emph{i)} generalize discrete diffusion steps to continuous diffusion steps and, then \emph{ii)} design a bijective mapping between continuous diffusion steps and continuous noise levels, where these noise levels indicate the amount of noise in data. Finally, we use this bijective mapping to construct an approximate diffusion process and an approximate reverse process, respectively.

\subsection{Bijective mapping between Continuous Diffusion Steps and Noise Levels}
\label{sec: map R and T}
In this section, we generalize discrete (integer) diffusion steps to continuous (real-valued) diffusion steps. Then, we introduce a bijective mapping $\gR$ and $\gT = R^{-1}$ between continuous diffusion steps $t$ and noise levels $r$: $r=\gR(t)$ and $t=\gT(r)$.
%
\paragraph{Define $\gR$.} 
We start with an integer diffusion step $t$. From \eqref{q(x_t|x_0)}, one can observe $x_t = \sqrt{\bar{\alpha}_t}\cdot x_0 + \sqrt{1-\bar{\alpha}_t}\cdot \epsilon$ where $\epsilon\sim\gN(0,I)$, thus sampling $x_t$ given $x_0$ is equivalent to adding a Gaussian noise to $x_0$.
Based on this observation, we define the noise level at step $t$ as $\gR(t)=\sqrt{\bar{\alpha}_t}$, which means $x_t$ is composed of $\gR(t)$ fraction of the data $x_0$ and $(1-\gR(t))$ fraction of white noise. For example, $\gR(t) = 0$ means no noise and $\gR(t) = 1$ means pure white noise. 
Next, we extend the domain of $\gR$ to real values. Assume that the variance schedule $\{\beta_t\}_{t=1}^T$ is linear: $\beta_i = \beta_1 + (i-1)\Delta\beta$, where $\Delta\beta=\frac{\beta_T-\beta_1}{T-1}$ \citep{ho2020denoising}. 
We further define an auxiliary constant $\hat{\beta}=\frac{1-\beta_1}{\Delta\beta}$, which is $\gg T$ assuming that $\beta_T\ll1.0$. \footnote{E.g., $\beta_1 = 1\times10^{-4}$, $\beta_T=0.02$ in \citet{ho2020denoising,kong2020diffwave}.} Then, we have
\begin{align}\label{eq: map R}
\begin{array}{rl}
    \bar{\alpha}_t & \displaystyle = \prod_{i=1}^t (1-\beta_i) 
    = \prod_{i=1}^t \left(1-\beta_1-(i-1)\Delta\beta \right)
    = (\Delta\beta)^t\prod_{i=0}^{t-1}\left(\hat{\beta}-i\right) \\
    & \displaystyle = (\Delta\beta)^t \Gamma\left(\hat{\beta}+1\right) \Gamma\left(\hat{\beta}-t+1\right)^{-1}.
\end{array}
\end{align}
Because the Gamma function $\Gamma$ is well-defined on $(0,\infty)$, \eqref{eq: map R} gives rise to a natural extension of $\bar{\alpha}_t$ for continuous diffusion steps $t$. As a result, for $t\in[0,\hat{\beta})$, we define the noise level at $t$ as:
\begin{equation}
    \gR(t) = (\Delta\beta)^{\frac t2}\Gamma\left(\hat{\beta}+1\right)^{\frac12} \Gamma\left(\hat{\beta}-t+1\right)^{-\frac12}.
\end{equation}

\paragraph{Define $\gT$.} For any noise level $r\in(0,1)$, its corresponding (continuous) diffusion step, $\gT(r)$, is defined by inverting $\gR$:
\begin{align}\label{eq: map T}
    \gT(r) = \gR^{-1}(r).
\end{align}
 By Stirling's approximation to Gamma functions, we have
\begin{align}\label{eq: log noise}
    2\log \gR(t) = t\Delta\beta + \left(\hat{\beta}+\frac12\right)\log\hat{\beta} 
    &- \left(\hat{\beta}-t+\frac12\right)\log(\hat{\beta}-t)-t \nonumber \\
    &+ \frac{1}{12}\left(\frac{1}{\hat{\beta}}-\frac{1}{\hat{\beta}-t}\right) +\gO(T^{-2})
\end{align}

Given a noise level $r=\gR(t)$, we numerically solve $t = \gT(r)$ by applying a binary search based on \eqref{eq: log noise}. We have $\gT(r)\in[t,t+1]$ for $r\in[\sqrt{\bar{\alpha}_{t+1}}, \sqrt{\bar{\alpha}_{t}}]$, and this provides a good initialization to the binary search algorithm. Experimentally, we find the binary search algorithm converges with high precision in no more than 20 iterations.

\subsection{Approximate the Diffusion Process}\label{sec: approx q}
Let $\hat{x}_0\sim\pdata$. Given a sequence of noise levels $1>r_1>r_2>\cdots>r_S>0$, we aim to construct each step in the approximate diffusion process as $\hat{x}_s\sim\gN(\hat{x}_s; r_s\hat{x}_0, (1-r_s^2)I)$. To achieve this goal, we define $\gamma_s = r_s^2/r_{s-1}^2$, compute the corresponding variances as $\eta_s = 1 - \gamma_s = 1-r_s^2/r_{s-1}^2$, and then define the transition probability in the approximate diffusion process as
\begin{equation}
    q(\hat{x}_s|\hat{x}_{s-1}) = \gN(\hat{x}_s; \sqrt{1-\eta_s}\hat{x}_{s-1}, \eta_sI) = \gN\left(\hat{x}_s; \frac{r_s}{r_{s-1}}\hat{x}_{s-1}, \left(1-\frac{r_s^2}{r_{s-1}^2}\right)I\right).
\end{equation}
One can see this by rewriting \eqref{q(x_t|x_0)}: $\eta_s$ corresponds to $\beta_t=1-\alpha_t$, $\gamma_s$ corresponds to $\alpha_t$, and $r_s$ corresponds to $\sqrt{\bar{\alpha}_t}$.
We then propose the following two ways to schedule the noise levels $\{r_s\}_{s=1}^S$. 

\paragraph{Noise levels from variances (VAR).} We start from the variance schedule $\{\eta_s\}_{s=1}^S$. Next, we compute $\gamma_s = 1 - \eta_s$ and $\bar{\gamma}_s = \prod_{i=1}^s\gamma_i$. The noise level at step $s$ is then $r_s = \sqrt{\bar{\gamma}_s}$.

\paragraph{Noise levels from steps (STEP).} We start from a subset of diffusion steps $\{\tau_s\}_{s-1}^S$ in $\{1,\cdots,T\}$. Then, the noise level at step $s$ is $r_s = \gR(\tau_s) = \sqrt{\bar{\alpha}_{\tau_s}}$. 

When $\eta_s=1-\bar{\alpha}_{\tau_s} / \bar{\alpha}_{\tau_{s-1}}$, we have $\bar{\gamma}_s = \bar{\alpha}_{\tau_s}$. Therefore, noise levels from steps can be regarded as a special case of noise levels from variances.

\subsection{Approximate the Reverse Process}\label{sec: approx p_theta}

Given the same sequence of noise levels in Section \ref{sec: approx q}, we aim to approximate the reverse process \eqref{eq: reverse process} in the original DDPM. To achieve this goal, we regard the model $\epsilon_{\theta}$ as being trained on variances $\{\eta_s\}_{s=1}^S$ instead of the original $\{\beta_t\}_{t=1}^T$. Then, the transition probability in the approximate reverse process is
\begin{equation}\label{eq: approx p_theta}
    p_{\theta}(\hat{x}_{s-1}|\hat{x}_s) = \gN\left(\hat{x}_{s-1}; \frac{1}{\sqrt{\gamma_s}}\left(\hat{x}_s-\frac{\eta_s}{\sqrt{1-\bar{\gamma}_s}}\epsilon_{\theta}(\hat{x}_s, \gT(r_s))\right), \tilde{\eta}_s I\right),
\end{equation}
where $\tilde{\eta}_s=\frac{1-\bar{\gamma}_{s-1}}{1-\bar{\gamma}_s}\eta_s$ for $s>1$ and $\tilde{\eta}_1=\eta_1$. $\tilde{\eta}_s$ corresponds to the $\tilde{\beta}_t=\sigma_t^2$ term.
There are two ways to sample from the approximate reverse process in \eqref{eq: approx p_theta}. Let every $\hat{\epsilon}_s$ be i.i.d. standard Gaussians for $1\leq s\leq S$.

\paragraph{DDPM reverse process (DDPM-rev).} The sampling procedure based on the DDPM reverse process is based on \eqref{eq: approx p_theta}: that is, to first sample $\hat{x}_S\sim\platent$ and then sample

\begin{equation}
\hat{x}_{s-1} = \frac{1}{\sqrt{\gamma_s}}\left(\hat{x}_s-\frac{\eta_s}{\sqrt{1-\bar{\gamma}_s}}\epsilon_{\theta}(\hat{x}_s, \gT(r_s))\right) + \sqrt{\tilde{\eta}_s}\hat{\epsilon}_s.
\end{equation}

\paragraph{DDIM reverse process (DDIM-rev).} Let $\kappa\in[0,1]$ be a hyperparameter. \footnote{$\kappa$ is $\eta$ in \citet{song2020denoising}.} Then, the sampling procedure based on DDIM \citep{song2020denoising} is to first sample $\hat{x}_S\sim\platent$ and then sample

\begin{equation}
    \hat{x}_{s-1} = \sqrt{\bar{\gamma}_{s-1}}\left(\frac{\hat{x}_s-\sqrt{1-\bar{\gamma}_s}\epsilon_{\theta}(\hat{x}_s, \gT(r_s))}{\sqrt{\bar{\gamma}_s}}\right) + \sqrt{1-\bar{\gamma}_{s-1}-\kappa^2\tilde{\eta}_s}\epsilon_{\theta}(\hat{x}_s, \gT(r_s)) + \kappa\sqrt{\tilde{\eta}_s}\hat{\epsilon}_s.
\end{equation}

When $\kappa=1$, the coefficient of the $\epsilon_{\theta}$ term in the DDIM reverse process is
\begin{equation}
    \begin{array}{rl}
        \displaystyle -\frac{\sqrt{1-\bar{\gamma}_s}}{\sqrt{\gamma_s}} + \sqrt{1-\bar{\gamma}_{s-1}-\frac{1-\bar{\gamma}_{s-1}}{1-\bar{\gamma}_{s}}\eta_s} 
        & \displaystyle = -\frac{1-\bar{\gamma}_s}{\sqrt{\gamma_s(1-\bar{\gamma}_s)}} + \frac{\sqrt{(\gamma_s-\bar{\gamma}_s)(1-\bar{\gamma}_s-\eta_s)}}{\sqrt{\gamma_s(1-\bar{\gamma}_s)}} \\
        & \displaystyle = -\frac{1-\bar{\gamma}_s}{\sqrt{\gamma_s(1-\bar{\gamma}_s)}} + \frac{\gamma_s-\bar{\gamma}_s}{\sqrt{\gamma_s(1-\bar{\gamma}_s)}} \\
        & \displaystyle = -\frac{\eta_s}{\sqrt{\gamma_s(1-\bar{\gamma}_s)}}.
    \end{array}
\end{equation}
Therefore, the DDPM reverse process is a special case of the DDIM reverse process ($\kappa=1$).

\subsection{Connections with Previous Methods}
The DDIM \citep{song2020denoising} method is equivalent to selecting noise levels from steps and using DDIM-rev in FastDPM. The fast sampling algorithm by DiffWave \citep{kong2020diffwave} is related to selecting noise levels from variances and using DDPM-rev in FastDPM. Compared with DiffWave, FastDPM offers an automatic way to select variances in different settings and a more natural way to compute noise levels.

\section{Experiments}
\label{sec:experiments}

In this section, we aim to answer the following two questions for FastDPM:
\begin{itemize}
    \item Which approximate diffusion process, VAR or STEP, is better?
    \item Which approximate reverse process, DDPM-rev or DDIM-rev, is better?
\end{itemize}
We investigate these questions by conducting extensive experiments in both image and audio domains.

\subsection{Setup}
\label{sec: exp setup}

\textbf{Image datasets.} We conduct unconditional image generation experiments on three datasets: CIFAR-10 (50k object images of resolution $32\times32$ \citep{krizhevsky2009learning}), CelebA ($\sim$163k face images of resolution $64\times64$ \citep{liu2015faceattributes}), and LSUN-bedroom ($\sim$3M bedroom images of resolution $256\times256$ \citep{yu2015lsun}). 

\textbf{Audio datasets.} We conduct unconditional and class-conditional audio synthesis experiments on the Speech Commands 0-9 (SC09) dataset, the spoken digit subset of the full Speech Commands dataset \citep{warden2018speech}. SC09 contains $\sim$31k one-second long utterances of ten classes (0 through 9) with a sampling rate of 16kHz. We conduct neural vocoding experiments (audio synthesis conditioned on mel spectrogram) on the LJSpeech dataset \citep{Ito2017ljspeech}. It contains $\sim$24 hours of audio ($\sim$13k utterances from a female speaker) recorded in home environment with a sampling rate of 22.05kHz.

\textbf{Models.} In all experiments, we use pretrained checkpoints in prior works. In detail, the pretrained models for CIFAR-10 and LSUN-bedroom are taken from DDPM \citep{ho2020denoising,Esser2020pytorch}, the pretrained model for CelebA is taken from DDIM \citep{song2020denoising}. In these models, $T$ is $1000$. The pretrained models for SC09 and LJSpeech are taken from DiffWave \citep{kong2020diffwave}. In these models, $T$ is $200$. In all models, $\beta_1=10^{-4}$, $\beta_T=2\times10^{-2}$, and all $\beta_t$'s are linearly interpolated between $\beta_1$ and $\beta_T$.

\textbf{Noise level schedules.} For each of the approximate diffusion process in Section \ref{sec: approx q}, we examine two schedules: linear and quadratic. 
For noise levels $\{\eta_s\}_{s=1}^S$ from variances, the two schedules are: 
\begin{itemize}
\vspace{-0.2em}
    \item Linear (VAR): $\eta_s = (1 + cs)~ \eta_0$.
    \item Quadratic (VAR): $\eta_s = (1 + cs)^2 ~ \eta_0$. 
\vspace{-0.2em}
\end{itemize}
We let $\eta_0=\beta_0$ and the constant $c$ satisfy $\prod_{s=1}^S(1-\eta_s)=\bar{\alpha}_T$. The noise level at step $s$ is $r_s=\sqrt{\bar{\gamma}_s}$. 

For noise levels $\{\eta_s\}_{s=1}^S$ from steps, they are computed from selected steps $\{\tau_s\}_{s=1}^S$ among $\{1,\cdots,T\}$ \citep{song2020denoising}. The two schedules are:
\begin{itemize}
\vspace{-0.2em}
    \item Linear (STEP): $\tau_s = \lfloor cs \rfloor$, where $c=\frac TS$.
    \item Quadratic (STEP): $\tau_s = \lfloor cs^2 \rfloor$, where $c=\frac45\cdot\frac {T}{S^2}$.
\vspace{-0.2em}
\end{itemize}
Then, the noise level at step $s$ is $r_s = \gR(\tau_s) = \sqrt{\bar{\alpha}_{\tau_s}}$. 

In image generation experiments, we follow the same noise level schedules as in \citet{song2020denoising}: quadratic schedules for CIFAR-10 and linear schedules for CelebA and LSUN-bedroom. We use linear schedules in SC09 experiments and quadratic schedules in LJSpeech experiments; we find these schedules have better quality.

\textbf{Evaluations.} In all unconditional generation experiments, we use the Fr{\'e}chet Inception Distance (FID) \citep{heusel2017gans, Lang2020fid} to evaluate generated samples. For the training set $X_t$ and the set of generated samples $X_g$, the FID between these two sets is defined as
\begin{equation}\label{eq: fid}
    \mathrm{FID} = \|\mu_t-\mu_g\|^2 + \mathrm{tr}\left(\Sigma_t+\Sigma_g-2\sqrt{\Sigma_t\Sigma_g}\right),
\end{equation}
where $\mu_{t}, \mu_{g}$ and $\Sigma_{t}, \Sigma_{g}$ are the means and covariances of $X_{t}, X_{g}$ after a feature transformation. In each image generation experiment, $X_g$ is 50K generated images. The transformed feature is the 2048-dimensional vector output of the last layer of Inception-V3 \citep{DBLP:journals/corr/SzegedyVISW15}. In each audio synthesis experiment, $X_g$ is 5K generated utterances. The transformed feature is the 1024-dimensional vector output of the last layer of a ResNeXT classifier \citep{ResNeXTcode}, which achieves $99.06\%$ accuracy on the training set and $98.76\%$ accuracy on the test set. The FID is the smaller the better. 

In the class-conditional generation experiment on SC09, we evaluate with accuracy and the Inception Score (IS) \citep{salimans2016improved}. \footnote{Note that FID is not an appropriate metric for conditional generation.} The accuracy is computed by matching the predictions of the ResNeXT classifier and the pre-specified labels in the dataset. The IS of generated samples $X_g$ is defined as
\begin{equation}\label{eq: is}
    \mathrm{IS} = \exp\left(\mathbb{E}_{x\sim X_g}\mathrm{KL}(p(x)\|\mathbb{E}_{x'\sim X_g}p(x'))\right),
\end{equation}
where $p(x)$ is the logit vector of the ResNeXT classifier. The IS and accuracy are the larger the better.

In the neural vocoding experiment on LJSpeech, we evaluate the speech quality with the crowdMOS tookit~\citep{ribeiro2011crowdmos}, where the test utterances from all models were presented to Mechanical Turk workers. We report the 5-scale Mean Opinion Scores~(MOS), and it is the larger the better.

\subsection{Results}

We report image generation results under different approximate diffusion processes, approximate reverse processes and $S$, the length of FastDPM. Evaluation results on CIFAR-10, CelebA, and LSUN-bedroom measured in FID are shown in Table \ref{tab: cifar fid}, Table \ref{tab: celeba fid}, and Table \ref{tab: lsun fid}, respectively. 

We report audio synthesis results under different approximate diffusion processes, approximate reverse processes and $S$, the length of FastDPM. Evaluation results of unconditional generation on SC09 measured in FID and IS are shown in Table \ref{tab: sc09 uncond}. Evaluation results of class-conditional generation on SC09 measured in accuracy and IS are shown in Table \ref{tab: sc09 cond}. Evaluation results of neural vocoding on LJSpeech measured in MOS are shown in Table \ref{tab: ljspeech}.

We display some generated samples of FastDPM, including image samples and mel-spectrogram of audio samples, in Appendix \ref{appendix: samples}. More audio samples can be found on the demo website. \footnote{Demo: \url{https://fastdpm.github.io}. Code: \url{https://github.com/FengNiMa/FastDPM_pytorch}}

\begin{table}[!t]
\vspace{-.3em}
    \centering
    \caption{CIFAR-10 image generation measured in FID. STEP means noise levels from steps and VAR means noise levels from variances. Both use quadratic schedules. $S$ is the length of FastDPM. The standard DDPM~($T=1000$) has $\mathrm{FID}=3.03$.}
    \vspace{0.3em}
    \label{tab: cifar fid}
    \begin{tabular}{cc|cccc}
        \Xhline{2\arrayrulewidth}
        Approx. & Approx. & \multicolumn{4}{c}{FID~($\downarrow$)} \\ \cline{3-6}
        Diffusion & Reverse & $S=10$ & $S=20$ & $S=50$ & $S=100$ \\ \Xhline{2\arrayrulewidth}
        STEP & DDIM-rev ($\kappa=0.0$)  & 11.01 &  \textbf{5.05} & \textbf{3.20} & \textbf{2.86} \\
        VAR  & DDIM-rev ($\kappa=0.0$)  &  \textbf{9.90} &  5.22 & 3.41 & 3.01 \\ \hline
        STEP & DDIM-rev ($\kappa=0.2$)  & 11.32 &  5.16 & 3.27 & 2.87 \\
        VAR  & DDIM-rev ($\kappa=0.2$)  & 10.18 &  5.32 & 3.50 & 3.04 \\ \hline
        STEP & DDIM-rev ($\kappa=0.5$)  & 13.53 &  6.14 & 3.61 & 3.05 \\
        VAR  & DDIM-rev ($\kappa=0.5$)  & 12.22 &  6.55 & 3.86 & 3.15 \\ \hline
        STEP & DDPM-rev                 & 36.70 & 14.82 & 5.79 & 4.03 \\
        VAR  & DDPM-rev                 & 29.43 & 15.27 & 6.74 & 4.58 \\ 
        \Xhline{2\arrayrulewidth}
    \end{tabular}
\end{table}

\begin{table}[!t]
\vspace{-.3em}
    \centering
    \caption{CelebA image generation measured in FID. STEP means noise levels from steps and VAR means noise levels from variances. Both use linear schedules. $S$ is the length of FastDPM. The standard DDPM~($T=1000$) has $\mathrm{FID}=7.00$.}
    \vspace{0.3em}
    \label{tab: celeba fid}
    \begin{tabular}{cc|cccc}
        \Xhline{2\arrayrulewidth}
        Approx. & Approx. & \multicolumn{4}{c}{FID~($\downarrow$)} \\ \cline{3-6}
        Diffusion & Reverse & $S=10$ & $S=20$ & $S=50$ & $S=100$ \\ \Xhline{2\arrayrulewidth}
        STEP & DDIM-rev ($\kappa=0.0$)  & 15.72 & 10.77 &  \textbf{8.31} &  \textbf{7.85} \\
        VAR  & DDIM-rev ($\kappa=0.0$)  & \textbf{15.31} & \textbf{10.69} &  8.41 &  7.95 \\ \hline
        STEP & DDPM-rev                 & 29.52 & 19.38 & 12.83 & 10.35 \\
        VAR  & DDPM-rev                 & 28.98 & 18.89 & 12.83 & 10.39 \\ 
        \Xhline{2\arrayrulewidth}
    \end{tabular}
\end{table}
\begin{table}[!t]
\vspace{-.3em}
    \centering
    \caption{LSUN-bedroom image generation measured in FID. STEP means noise levels from steps and VAR means noise levels from variances. Both use linear schedules. $S$ is the length of FastDPM.}
    \vspace{0.3em}
    \label{tab: lsun fid}
    \begin{tabular}{cc|cccc}
        \Xhline{2\arrayrulewidth}
        Approx. & Approx. & \multicolumn{4}{c}{FID~($\downarrow$)} \\ \cline{3-6}
        Diffusion & Reverse & $S=10$ & $S=20$ & $S=50$ & $S=100$ \\ \Xhline{2\arrayrulewidth}
        STEP & DDIM-rev ($\kappa=0.0$)  & \textbf{19.07} &  9.95 &  8.43 &  9.94 \\
        VAR  & DDIM-rev ($\kappa=0.0$)  & 19.98 &  \textbf{9.86} &  \textbf{8.37} & 10.27 \\ \hline
        STEP & DDPM-rev                 & 42.69 & 20.97 & 10.24 &  \textbf{7.98} \\
        VAR  & DDPM-rev                 & 41.00 & 20.12 & 10.12 &  8.13 \\ 
        \Xhline{2\arrayrulewidth}
    \end{tabular}
\end{table}

\begin{table}[!t]
\vspace{-.3em}
    \centering
    \caption{SC09 unconditional audio synthesis measured in FID and IS. STEP means noise levels from steps and VAR means noise levels from variances. Both use linear schedules. $S$ is the length of FastDPM. The original DiffWave~($T=200$) has $\mathrm{FID}=1.29$ and IS$=5.30$.}
    \vspace{0.3em}
    \label{tab: sc09 uncond}
    \begin{tabular}{cc|ccc|ccc}
        \Xhline{2\arrayrulewidth}
        Approx. & Approx. & \multicolumn{3}{c|}{FID~($\downarrow$)} & \multicolumn{3}{c}{IS~($\uparrow$)} \\ \cline{3-8}
        Diffusion & Reverse & $S=10$ & $S=20$ & $S=50$ & $S=10$ & $S=20$ & $S=50$ \\ \Xhline{2\arrayrulewidth}
        STEP & DDIM-rev ($\kappa=0.0$)  & 4.72 & 5.31 & 5.54 & 2.46 & 2.27 & 2.23 \\
        VAR  & DDIM-rev ($\kappa=0.0$)  & 4.74 & 4.88 & 5.58 & 2.49 & 2.42 & 2.21 \\ \hline
        STEP & DDIM-rev ($\kappa=0.5$)  & 2.60 & 2.52 & 2.46 & 3.94 & 4.17 & 4.19 \\
        VAR  & DDIM-rev ($\kappa=0.5$)  & 2.67 & 2.49 & 2.47 & 3.94 & 4.20 & 4.20 \\ \hline
        STEP & DDPM-rev                 & 1.75 & 1.40 & \textbf{1.33} & 4.03 & 4.57 & 5.16 \\
        VAR  & DDPM-rev                 & \textbf{1.69} & \textbf{1.38} & 1.34 & \textbf{4.06} & \textbf{4.63} & \textbf{5.18} \\ 
        \Xhline{2\arrayrulewidth}
    \end{tabular}
\end{table}
\begin{table}[!t]
\vspace{-.3em}
    \centering
    \caption{SC09 class-conditional audio synthesis. The results are measured by accuracy and IS. STEP means noise levels from steps and VAR means noise levels from variances. Both use linear schedules. $S$ is the length of FastDPM. The DiffWave~($T=200$) has accuracy $=91.2\%$ and $\mathrm{IS}=6.63$.}
    \vspace{0.3em}
    \label{tab: sc09 cond}
    \begin{tabular}{cc|ccc|ccc}
        \Xhline{2\arrayrulewidth}
        Approx. & Approx. & \multicolumn{3}{c|}{Accuracy~($\uparrow$)} & \multicolumn{3}{c}{IS~($\uparrow$)} \\ \cline{3-8}
        Diffusion & Reverse & $S=10$ & $S=20$ & $S=50$ & $S=10$ & $S=20$ & $S=50$ \\ \Xhline{2\arrayrulewidth}
        STEP & DDIM-rev ($\kappa=0.0$)  & $66.5\%$ & $68.3\%$ & $66.1\%$ & $3.21$ & $3.18$ & $2.87$\\
        VAR  & DDIM-rev ($\kappa=0.0$)  & $66.6\%$ & $68.5\%$ & $66.1\%$ & $3.26$ & $3.22$ & $2.88$\\ \hline
        STEP & DDIM-rev ($\kappa=0.5$)  & $85.8\%$ & $\bf 88.4\%$ & $87.8\%$ & $\bf 5.79$ & $6.23$ & $6.00$ \\
        VAR  & DDIM-rev ($\kappa=0.5$)  & $\bf 86.0\%$ & $88.2\%$ & $\bf 88.0\%$ & $5.74$ & $\bf 6.24$ & $\bf 6.03$ \\ \hline
        STEP & DDPM-rev                 & $79.9\%$ & $82.7\%$ & $86.8\%$ & $4.71$ & $5.10$ & $5.83$\\
        VAR  & DDPM-rev                 & $81.0\%$ & $82.8\%$ & $87.0\%$ & $4.93$ & $5.16$ & $5.86$\\ 
        \Xhline{2\arrayrulewidth}
    \end{tabular}
\end{table}
\begin{table}[!h]
\vspace{-.3em}
    \centering
    \caption{LJSpeech audio synthesis conditioned on mel spectrogram measured. The results are measured by 5-scale MOS with 95\% confidence intervals. STEP means noise levels from steps and VAR means noise levels from variances. Both use quadratic schedules. $S$ is the length of FastDPM.}
    \vspace{0.3em}
    \label{tab: ljspeech}
    \begin{tabular}{ccc|c}
        \Xhline{2\arrayrulewidth}
        Approx. Diffusion & Approx. Reverse & $S$ & MOS~($\uparrow$) \\ \Xhline{2\arrayrulewidth}
        STEP & DDIM-rev ($\kappa=0.0$)  & 5 & $3.72\pm0.11$ \\
        VAR  & DDIM-rev ($\kappa=0.0$)  & 5 & $3.75\pm0.10$ \\ \hline
        STEP & DDPM-rev                 & 5 & $4.28\pm0.08$ \\
        VAR  & DDPM-rev                 & 5 & $\bf 4.31\pm0.07$ \\ \Xhline{2\arrayrulewidth}
        \multicolumn{2}{c}{DiffWave ($T=200$)} & 200 & $4.42\pm0.10$ \\
        \multicolumn{2}{c}{Ground truth} & -- & $4.51\pm0.07$ \\
        \Xhline{2\arrayrulewidth}
    \end{tabular}
\end{table}

\subsection{Observations and Insights}
We have the following observations and insights according to the above experimental results. 

\textbf{VAR marginally outperforms STEP for small $S$.} 
In the above experiments, the two approximate diffusion processes (STEP and VAR) generally match performances of each other. On CIFAR-10, VAR outperforms STEP when $S=10$, and STEP slightly outperforms VAR when $S\geq20$. On CelebA, VAR slightly outperforms STEP when $S\leq20$, and they have similar results when $S\geq50$. On LSUN-bedroom, VAR slightly outperforms STEP when $S\leq50$, and STEP slightly outperforms VAR when $S=100$. On SC09, VAR slightly outperforms STEP in most cases. On LJSpeech, VAR slightly outperforms STEP when $S=5$. Based on these results, we conclude that VAR marginally outperforms STEP for small $S$. 

\begin{figure}[!h]
    \centering
    \subfloat[][STEP + DDIM-rev ($\kappa=0.0$) (top) / DDIM-rev ($\kappa=0.5$) (middle) / DDPM-rev (bottom)]{
        \includegraphics[trim=150 40 270 60, clip, width=0.85\textwidth]{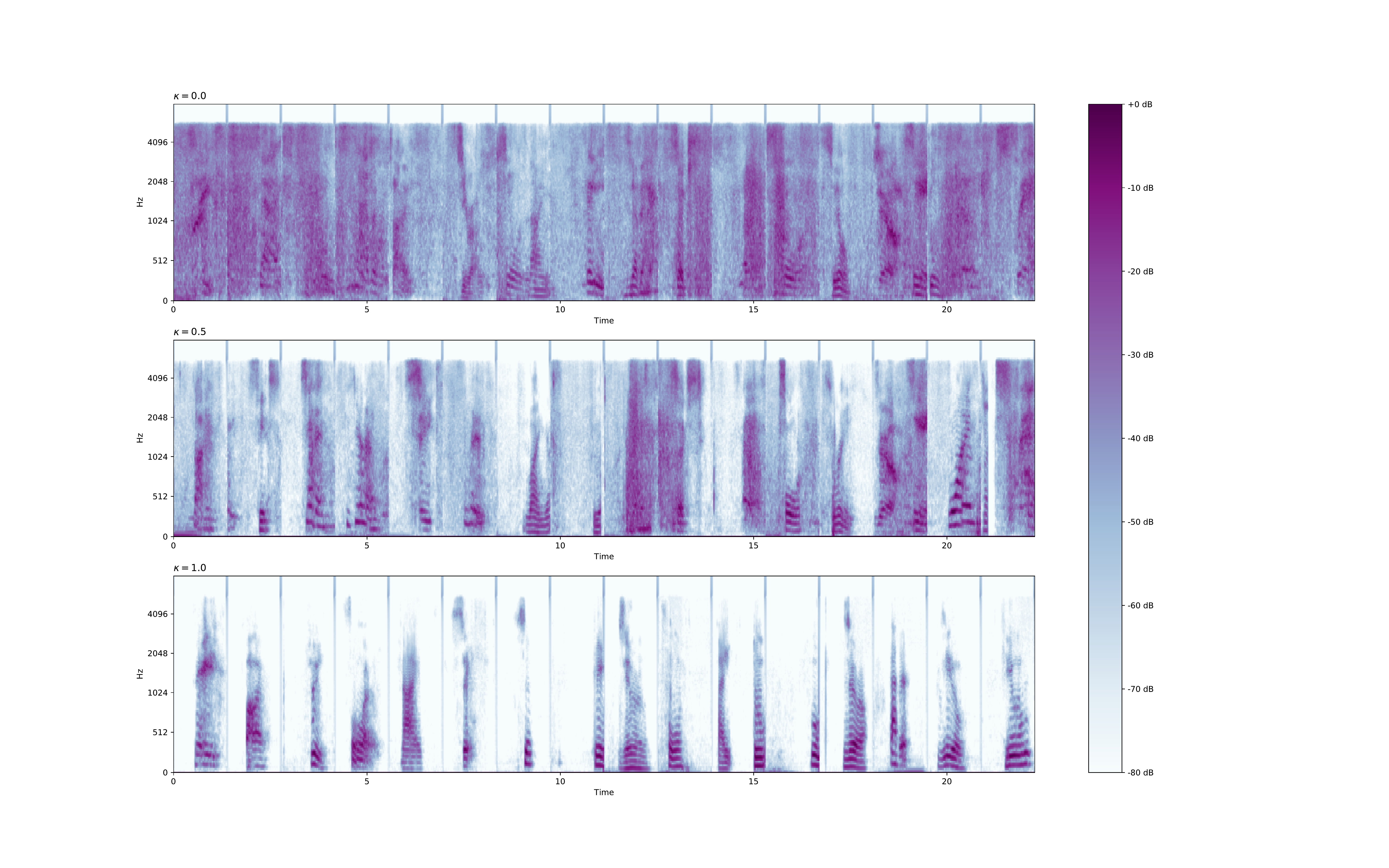}
    }\\
    \subfloat[][VAR + DDIM-rev ($\kappa=0.0$) (top) / DDIM-rev ($\kappa=0.5$) (middle) / DDPM-rev (bottom)]{
        \includegraphics[trim=150 40 270 60, clip, width=0.85\textwidth]{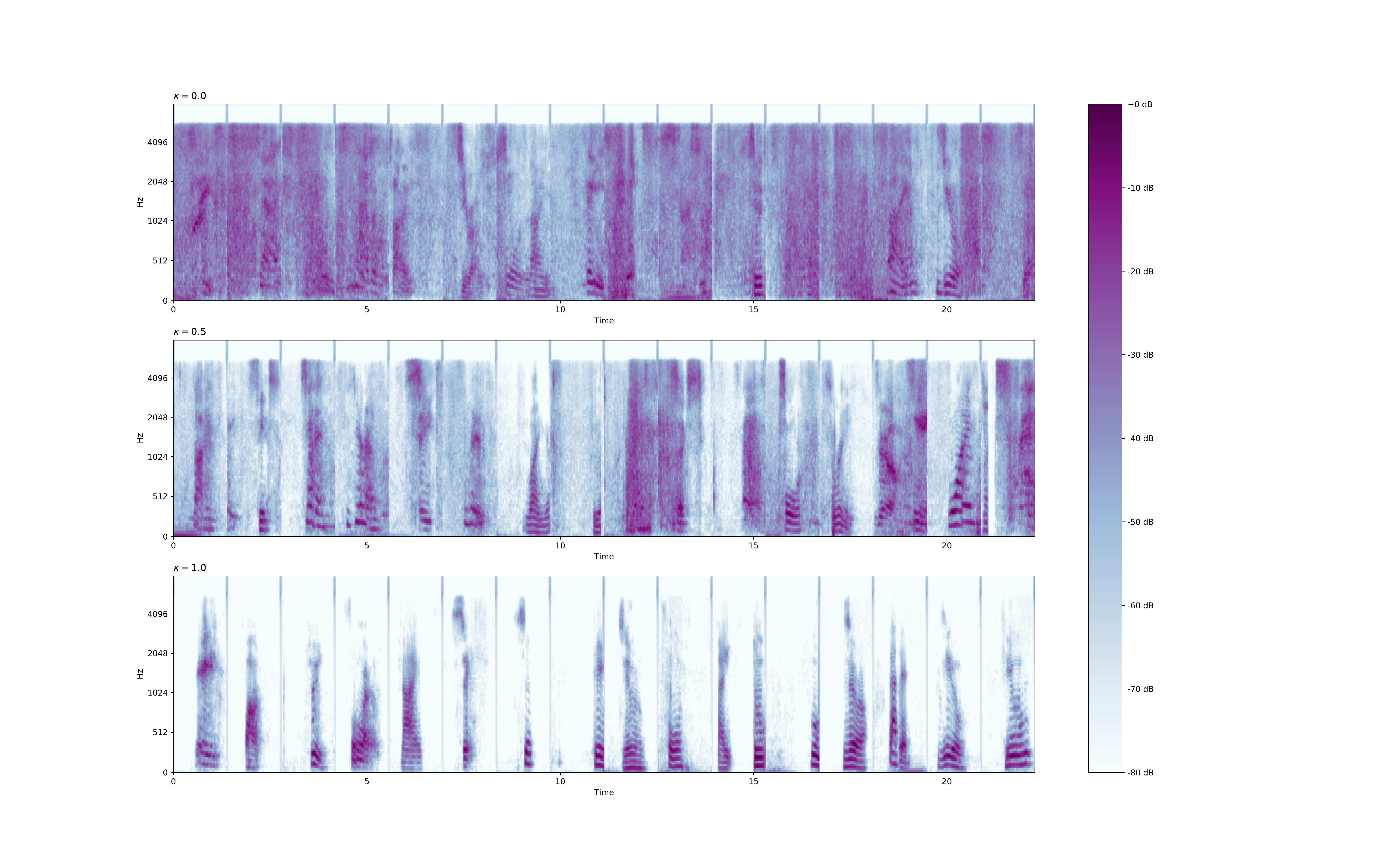}
    }
    \caption{Mel-spectrogram of 16 synthesized utterances ($S=50$). We use linear noise level schedules from steps in (a) and variances in (b). In each subplot, the top row shows results of DDIM-rev ($\kappa=0.0$), the middle row shows results of DDIM-rev ($\kappa=0.5$), and the bottom row shows results of DDPM-rev. DDPM-rev produces the clearest utterances in these approximate reverse processes.}
    \label{fig: sc09 uncond mel}
\end{figure}

\textbf{Different reverse processes dominate in different domains.} 
In the above experiments, the difference between DDPM and DDIM reverse processes is very clear. In image generation tasks, DDIM-rev significantly outperforms DDPM-rev except for the $S=100$ case in the LSUN-bedroom experiment. When we \textit{reduce} $\kappa$ from $1.0$ to $0.0$ (see Table \ref{tab: cifar fid}), the quality of generated samples consistently improves.
In contrast, in audio synthesis tasks, DDPM-rev significantly outperforms DDIM-rev. When we \textit{increase} $\kappa$ from $0.0$ to $1.0$ (see Table \ref{tab: sc09 uncond}), the quality of generated samples consistently improves. This can also be observed from Figure \ref{fig: sc09 uncond mel}: DDIM produces very noisy utterances while DDPM produces very clean utterances. 

The results indicate that in the image domain, DDIM-rev produces better quality whereas in the audio domain, DDPM-rev produces better quality.
We speculate the reason behind the difference is that in the audio domain, waveforms naturally exhibit significant amount of stochasticity. The DDPM reverse process offers much stochasticity because at each reverse step $s$, $\hat{x}_{s-1}$ is sampled from a Gaussian distribution. However, the DDIM reverse process ($\kappa=0.0$) is a deterministic mapping from latents to data, so it leads to degrade quality in the audio domain. 
This hypothesis is also aligned with previous result that the flow-based model with deterministic mapping was unable to generate intelligible speech unconditionally on SC09~\citep{ping2021waveflow_sc09}.

\textbf{The amount of conditional information affects the choice of reverse processes.}
In audio synthesis experiments, we find the amount of conditional information affects the generation quality of FastDPM with different reverse processes. In the unconditional generation experiment on SC09, DDPM-rev (which corresponds to $\kappa=1.0$) has the best results. When there is slightly more conditional information in the class-conditional generation experiment on SC09, DDIM-rev with $\kappa=0.5$ has the best results and slightly outperforms DDPM-rev. In both experiments DDIM-rev with $\kappa=0.0$ has much worse results. When there is much more conditional information~(mel spectrogram) in the neural vocoding experiments on LJSpeech, DDPM-rev is still better than DDIM-rev, but the difference between these two methods is reduced. 
%
We speculate that adding conditional information reduces the amount of stochasticity required. When there is no conditional information, we need a large amount of stochasticity ($\kappa=1.0$); when there is weak class information, we need moderate stochasticity ($\kappa=0.5$); and when there is strong mel-spectrogram information, even having no stochasticity ($\kappa=0.0$) is able to generate reasonable samples.

\section{Conclusion}
\label{sec:conclusion}
Diffusion models are a class of powerful deep generative models that produce superior quality samples on various generation tasks. In this paper, we introduce FastDPM, a unified framework for fast sampling in diffusion models without retraining. FastDPM generalizes prior methods and provides more flexibility. We extensively evaluate and analyze FastDPM in image and audio generation tasks.
One limitation of FastDPM is that when $S$ is small, there is still quality degradation compared to the original DDPM. We plan to study algorithms offering higher quality for extremely small $S$ in future.

\newpage
\bibliographystyle{abbrvnat}
\bibliography{main}

\newpage
\appendix

\section{Derivations for Diffusion Model}
\label{appendix:derivations}

\subsection{Derivation of $q(x_t|x_0)$}
According to the definition of diffusion process, we have
\begin{equation}
    x_t = \sqrt{\alpha_t}x_{t-1}+\sqrt{\beta_t}\epsilon_t,
\end{equation}
where each $\epsilon_t$ is an i.i.d. standard Gaussian. Then, by recursion, we have
\begin{equation}\begin{array}{rl}
    x_t & = \sqrt{\alpha_t\alpha_{t-1}}x_{t-2}+\sqrt{\alpha_t\beta_{t-1}}\epsilon_{t-1}+\sqrt{\beta_t}\epsilon_t \\
     & = \sqrt{\alpha_t\alpha_{t-1}\alpha_{t-1}}x_{t-3}+\sqrt{\alpha_t\alpha_{t-1}\beta_{t-2}}\epsilon_{t-2}+\sqrt{\alpha_t\beta_{t-1}}\epsilon_{t-1}+\sqrt{\beta_t}\epsilon_t \\
     & ~\vdots \\
     & = \sqrt{\bar{\alpha}_t}x_0 + \sqrt{\alpha_t\alpha_{t-1}\cdots\alpha_2\beta_1}\epsilon_1+\cdots+\sqrt{\alpha_t\beta_{t-1}}\epsilon_{t-1}+\sqrt{\beta_t}\epsilon_t.
\end{array}\end{equation}
As a result, $q(x_t|x_0)$ is still Gaussian. Its mean vector is $\sqrt{\bar{\alpha}_t}x_0$, and its covariance matrix is $(\alpha_t\alpha_{t-1}\cdots\alpha_2\beta_1+\cdots+\alpha_t\beta_{t-1}+\beta_t) I = (1-\bar{\alpha}_t) I$. Formally, we have
\begin{align}
q(x_t|x_0)=\gN(x_t;~ \sqrt{\bar{\alpha}_t}x_0,~ (1-\bar{\alpha}_t)  I).
\end{align}

\newpage

\section{Generated Samples in Experiments}\label{appendix: samples}

\subsection{Unconditional Generation on CIFAR-10}

\begin{figure}[!h]
    \centering
    \begin{tikzpicture}
	    \node (image) at (0,0) {
            \includegraphics[width=0.4\textwidth]{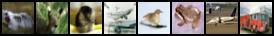}};
        \node (image) at (0,1) {
            \includegraphics[width=0.4\textwidth]{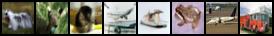}};
        \node (image) at (0,2) {
            \includegraphics[width=0.4\textwidth]{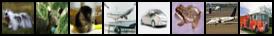}};
        \node (image) at (0,3) {
            \includegraphics[width=0.4\textwidth]{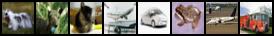}};
        \node (image) at (6,0) {
            \includegraphics[width=0.4\textwidth]{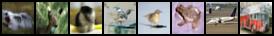}};
        \node (image) at (6,1) {
            \includegraphics[width=0.4\textwidth]{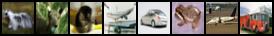}};
        \node (image) at (6,2) {
            \includegraphics[width=0.4\textwidth]{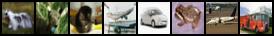}};
        \node (image) at (6,3) {
            \includegraphics[width=0.4\textwidth]{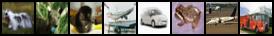}};
        \node[] at (-3.5,0) {$S=10$};
        \node[] at (-3.5,1) {$S=20$};
        \node[] at (-3.5,2) {$S=50$};
        \node[] at (-3.5,3) {$S=100$};
        \node[] at (0,3.7) {STEP};
        \node[] at (6,3.7) {VAR};
    \end{tikzpicture}
    \caption{Comparison of generated samples of FastDPM on CIFAR-10 among different $S$ and approximate diffusion processes. The approximate reverse process is DDIM-rev ($\kappa=0.0$).}
    \label{fig: cifar10 fix rev}
\end{figure}

\begin{figure}[!h]
    \centering
    \begin{tikzpicture}
	    \node (image) at (0,0) {
            \includegraphics[width=0.4\textwidth]{cifar10_VAR_S10_kappa0.0.jpg}};
        \node (image) at (0,1) {
            \includegraphics[width=0.4\textwidth]{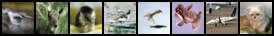}};
        \node (image) at (0,2) {
            \includegraphics[width=0.4\textwidth]{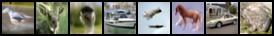}};
        \node (image) at (0,3) {
            \includegraphics[width=0.4\textwidth]{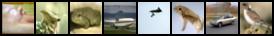}};
        \node (image) at (6,0) {
            \includegraphics[width=0.4\textwidth]{cifar10_VAR_S20_kappa0.0.jpg}};
        \node (image) at (6,1) {
            \includegraphics[width=0.4\textwidth]{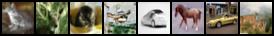}};
        \node (image) at (6,2) {
            \includegraphics[width=0.4\textwidth]{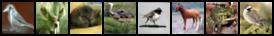}};
        \node (image) at (6,3) {
            \includegraphics[width=0.4\textwidth]{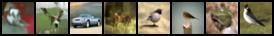}};
        \node[] at (-3.5,0) {$\kappa=0.0$};
        \node[] at (-3.5,1) {$\kappa=0.2$};
        \node[] at (-3.5,2) {$\kappa=0.5$};
        \node[] at (-3.7,3) {DDPM-rev};
        \node[] at (0,3.7)  {$S=10$};
        \node[] at (6,3.7)  {$S=20$};
        
        \node (image) at (0,-5) {
            \includegraphics[width=0.4\textwidth]{cifar10_VAR_S50_kappa0.0.jpg}};
        \node (image) at (0,-4) {
            \includegraphics[width=0.4\textwidth]{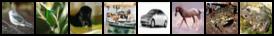}};
        \node (image) at (0,-3) {
            \includegraphics[width=0.4\textwidth]{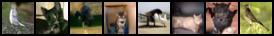}};
        \node (image) at (0,-2) {
            \includegraphics[width=0.4\textwidth]{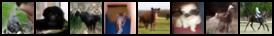}};
        \node (image) at (6,-5) {
            \includegraphics[width=0.4\textwidth]{cifar10_VAR_S100_kappa0.0.jpg}};
        \node (image) at (6,-4) {
            \includegraphics[width=0.4\textwidth]{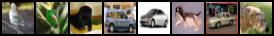}};
        \node (image) at (6,-3) {
            \includegraphics[width=0.4\textwidth]{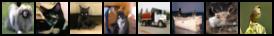}};
        \node (image) at (6,-2) {
            \includegraphics[width=0.4\textwidth]{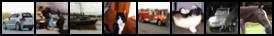}};
        \node[] at (-3.5,-5) {$\kappa=0.0$};
        \node[] at (-3.5,-4) {$\kappa=0.2$};
        \node[] at (-3.5,-3) {$\kappa=0.5$};
        \node[] at (-3.7,-2) {DDPM-rev};
        \node[] at (0,-1.3)  {$S=50$};
        \node[] at (6,-1.3)  {$S=100$};
    \end{tikzpicture}
    \caption{Comparison of generated samples of FastDPM on CIFAR-10 among different $S$ and approximate reverse processes. The approximate diffusion process is VAR.}
    \label{fig: cifar10 fix diff}
\end{figure}

\newpage
\subsection{Unconditional Generation on CelebA}
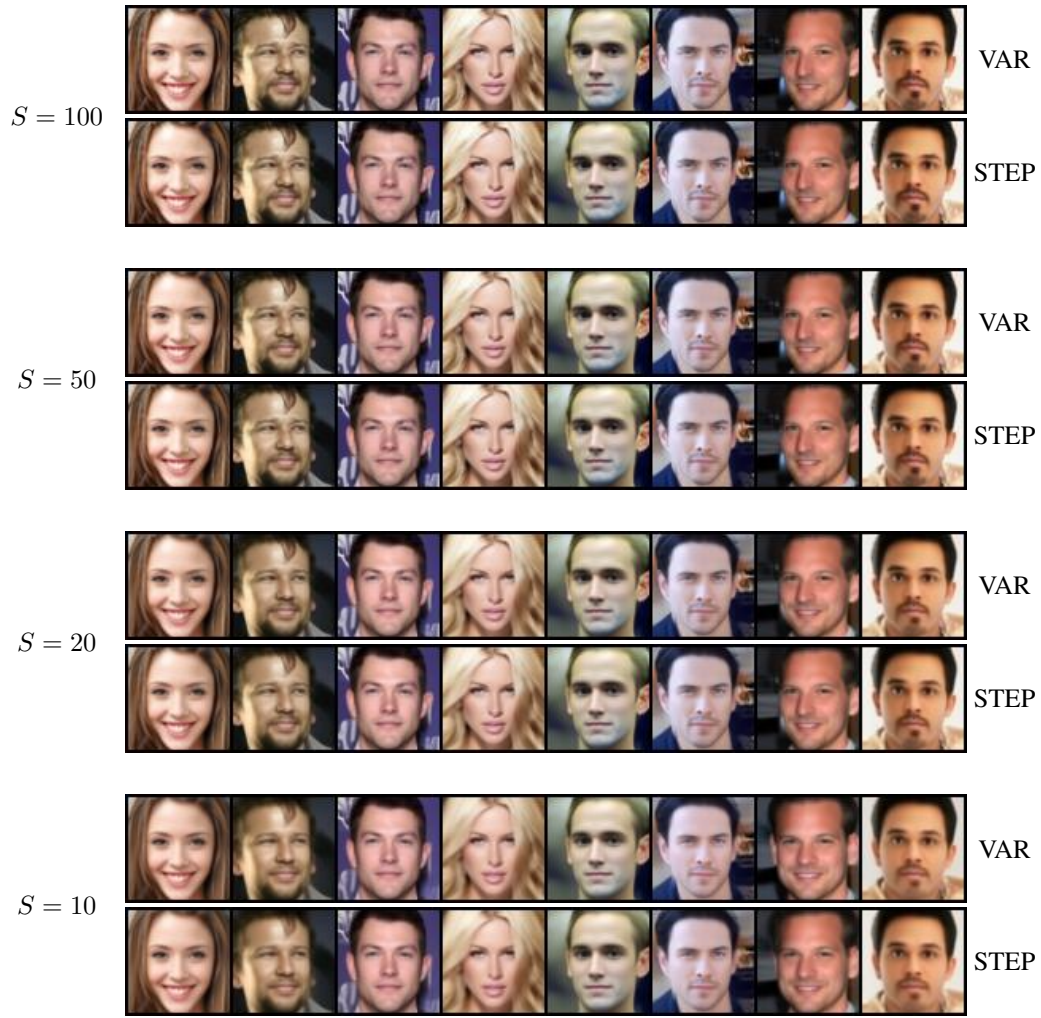
\begin{figure}[!h]
    \centering
    \begin{tikzpicture}
	    \node (image) at (0,0) {
            \includegraphics[width=0.8\textwidth]{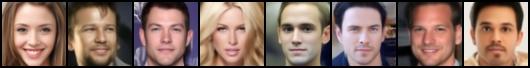}};
        \node (image) at (0,3.5) {
            \includegraphics[width=0.8\textwidth]{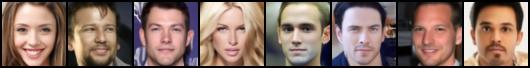}};
        \node (image) at (0,7) {
            \includegraphics[width=0.8\textwidth]{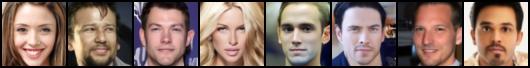}};
        \node (image) at (0,10.5) {
            \includegraphics[width=0.8\textwidth]{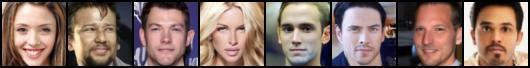}};
        \node (image) at (0,1.5) {
            \includegraphics[width=0.8\textwidth]{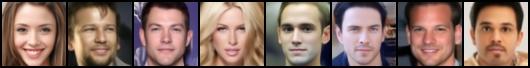}};
        \node (image) at (0,5) {
            \includegraphics[width=0.8\textwidth]{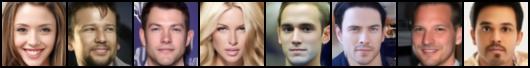}};
        \node (image) at (0,8.5) {
            \includegraphics[width=0.8\textwidth]{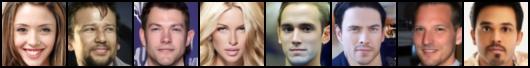}};
        \node (image) at (0,12) {
            \includegraphics[width=0.8\textwidth]{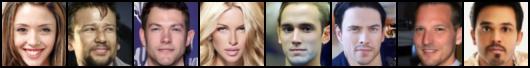}};
        \node[] at (-6.5,0.75) {$S=10$};
        \node[] at (-6.5,4.25) {$S=20$};
        \node[] at (-6.5,7.75) {$S=50$};
        \node[] at (-6.5,11.25) {$S=100$};
        \node[] at (6.1,0) {STEP};
        \node[] at (6.1,1.5) {VAR};
        \node[] at (6.1,3.5) {STEP};
        \node[] at (6.1,5) {VAR};
        \node[] at (6.1,7) {STEP};
        \node[] at (6.1,8.5) {VAR};
        \node[] at (6.1,10.5) {STEP};
        \node[] at (6.1,12) {VAR};
    \end{tikzpicture}
    \caption{Comparison of generated samples of FastDPM on CelebA among different $S$ and approximate diffusion processes. The approximate reverse process is DDIM-rev ($\kappa=0.0$).}
    \label{fig: celeba64 fix rev}
\end{figure}

\newpage
\subsection{Unconditional Generation on LSUN-bedroom}

\begin{figure}[!h]
    \centering
    \begin{tikzpicture}
	    \node (image) at (0,0) {
            \includegraphics[trim=0 0 1032 0, clip, height=3cm]{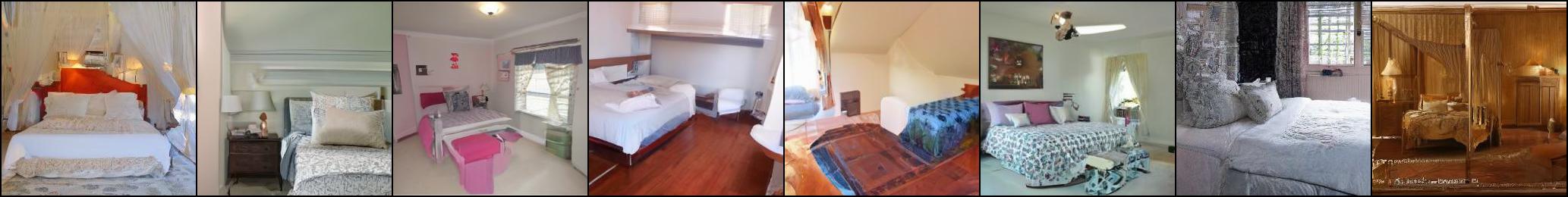}};
        \node (image) at (0,3) {
            \includegraphics[trim=1032 0 0 0, clip, height=3cm]{lsun_bedroom_STEP_S100_kappa0.0.jpg}};
        \node (image) at (0,6.5) {
            \includegraphics[trim=0 0 1032 0, clip, height=3cm]{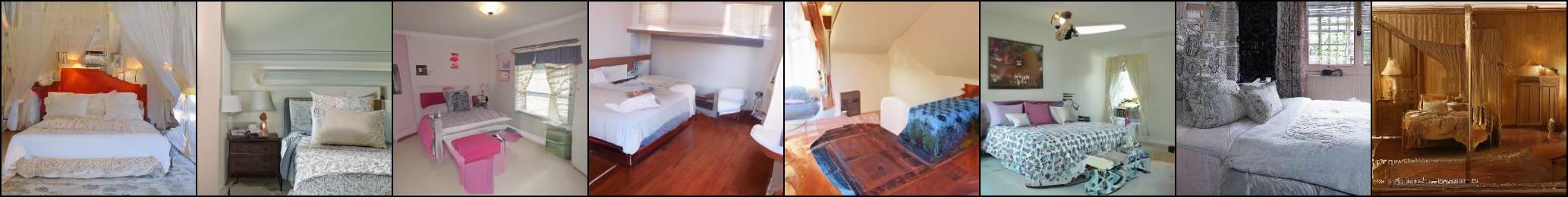}};
        \node (image) at (0,9.5) {
            \includegraphics[trim=1032 0 0 0, clip, height=3cm]{lsun_bedroom_VAR_S100_kappa0.0.jpg}};
        \node[] at (-6.75,1.5) {STEP};
        \node[] at (-6.75,8) {VAR};
    \end{tikzpicture}
    \caption{Comparison of generated samples of FastDPM on LSUN bedroom among different approximate diffusion processes. The approximate reverse process is DDIM-rev ($\kappa=0.0$) and $S=100$.}
    \label{fig: lsun fix rev-100}
\end{figure}

\newpage
\begin{figure}[!h]
    \centering
    \begin{tikzpicture}
	    \node (image) at (0,0) {
            \includegraphics[trim=0 0 1032 0, clip, height=3cm]{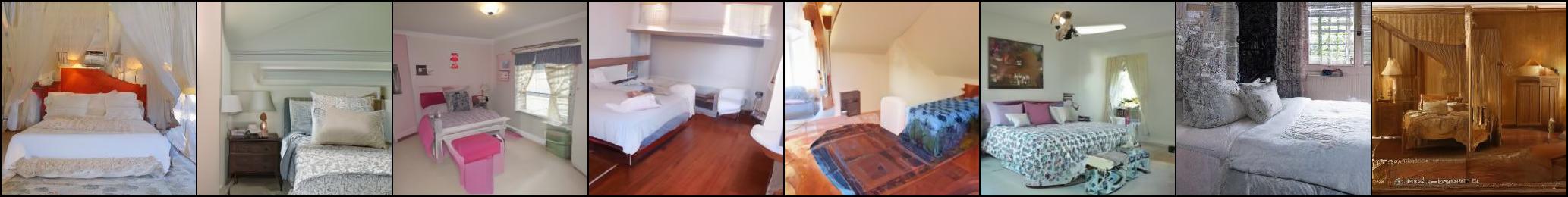}};
        \node (image) at (0,3) {
            \includegraphics[trim=1032 0 0 0, clip, height=3cm]{lsun_bedroom_STEP_S50_kappa0.0.jpg}};
        \node (image) at (0,6.5) {
            \includegraphics[trim=0 0 1032 0, clip, height=3cm]{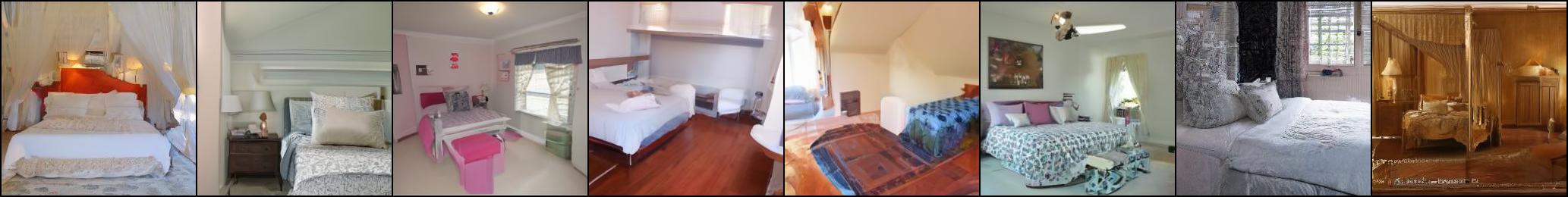}};
        \node (image) at (0,9.5) {
            \includegraphics[trim=1032 0 0 0, clip, height=3cm]{lsun_bedroom_VAR_S50_kappa0.0.jpg}};
        \node[] at (-6.75,1.5) {STEP};
        \node[] at (-6.75,8) {VAR};
    \end{tikzpicture}
    \caption{Comparison of generated samples of FastDPM on LSUN bedroom among different approximate diffusion processes. The approximate reverse process is DDIM-rev ($\kappa=0.0$) and $S=50$.}
    \label{fig: lsun fix rev-50}
\end{figure}

\newpage
\begin{figure}[!h]
    \centering
    \begin{tikzpicture}
	    \node (image) at (0,0) {
            \includegraphics[trim=0 0 1032 0, clip, height=3cm]{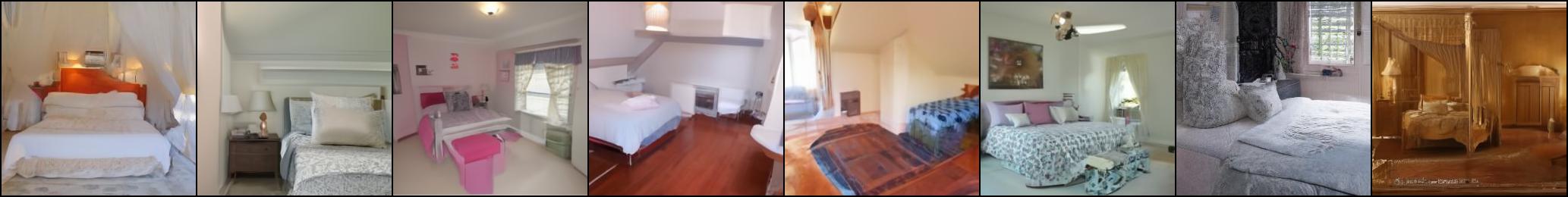}};
        \node (image) at (0,3) {
            \includegraphics[trim=1032 0 0 0, clip, height=3cm]{lsun_bedroom_STEP_S20_kappa0.0.jpg}};
        \node (image) at (0,6.5) {
            \includegraphics[trim=0 0 1032 0, clip, height=3cm]{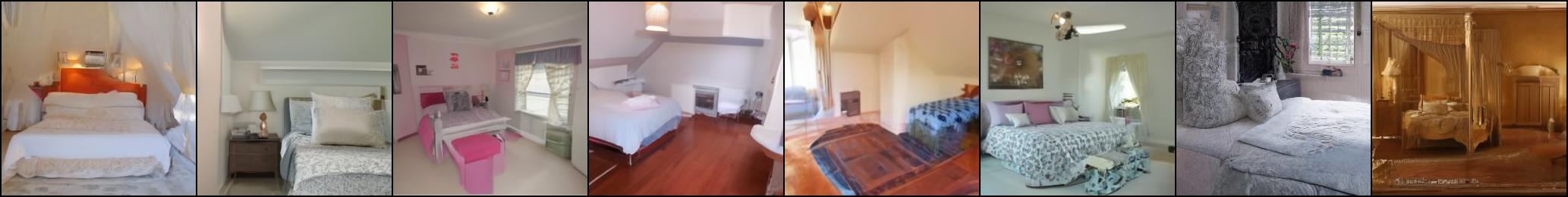}};
        \node (image) at (0,9.5) {
            \includegraphics[trim=1032 0 0 0, clip, height=3cm]{lsun_bedroom_VAR_S20_kappa0.0.jpg}};
        \node[] at (-6.75,1.5) {STEP};
        \node[] at (-6.75,8) {VAR};
    \end{tikzpicture}
    \caption{Comparison of generated samples of FastDPM on LSUN bedroom among different approximate diffusion processes. The approximate reverse process is DDIM-rev ($\kappa=0.0$) and $S=20$.}
    \label{fig: lsun fix rev-20}
\end{figure}

\newpage
\begin{figure}[!h]
    \centering
    \begin{tikzpicture}
	    \node (image) at (0,0) {
            \includegraphics[trim=0 0 1032 0, clip, height=3cm]{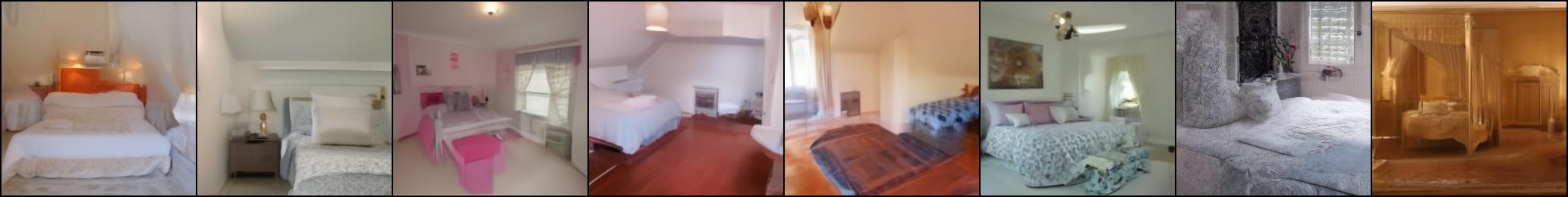}};
        \node (image) at (0,3) {
            \includegraphics[trim=1032 0 0 0, clip, height=3cm]{lsun_bedroom_STEP_S10_kappa0.0.jpg}};
        \node (image) at (0,6.5) {
            \includegraphics[trim=0 0 1032 0, clip, height=3cm]{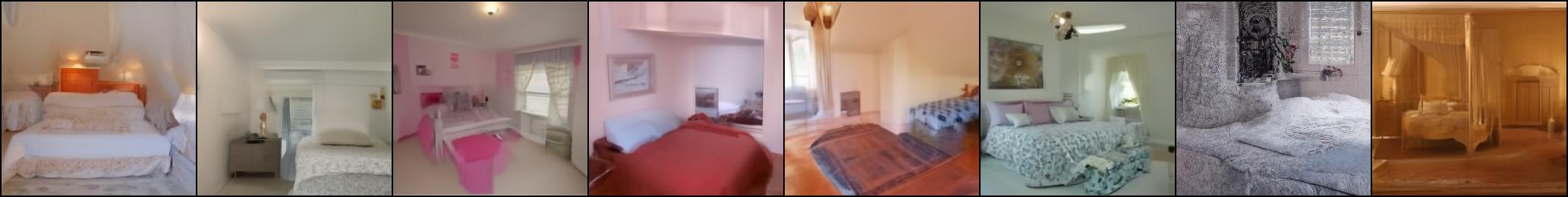}};
        \node (image) at (0,9.5) {
            \includegraphics[trim=1032 0 0 0, clip, height=3cm]{lsun_bedroom_VAR_S10_kappa0.0.jpg}};
        \node[] at (-6.75,1.5) {STEP};
        \node[] at (-6.75,8) {VAR};
    \end{tikzpicture}
    \caption{Comparison of generated samples of FastDPM on LSUN bedroom among different approximate diffusion processes. The approximate reverse process is DDIM-rev ($\kappa=0.0$) and $S=10$.}
    \label{fig: lsun fix rev-10}
\end{figure}

\newpage
\subsection{Unconditional Generation on SC09}

\newpage
\subsection{Conditional Generation on SC09}

\begin{figure}[!h]
    \centering
    \subfloat[][STEP + DDIM-rev ($\kappa=0.0$) (top) / DDIM-rev ($\kappa=0.5$) (middle) / DDPM-rev (bottom)]{
        \includegraphics[trim=150 40 270 60, clip, width=0.85\textwidth]{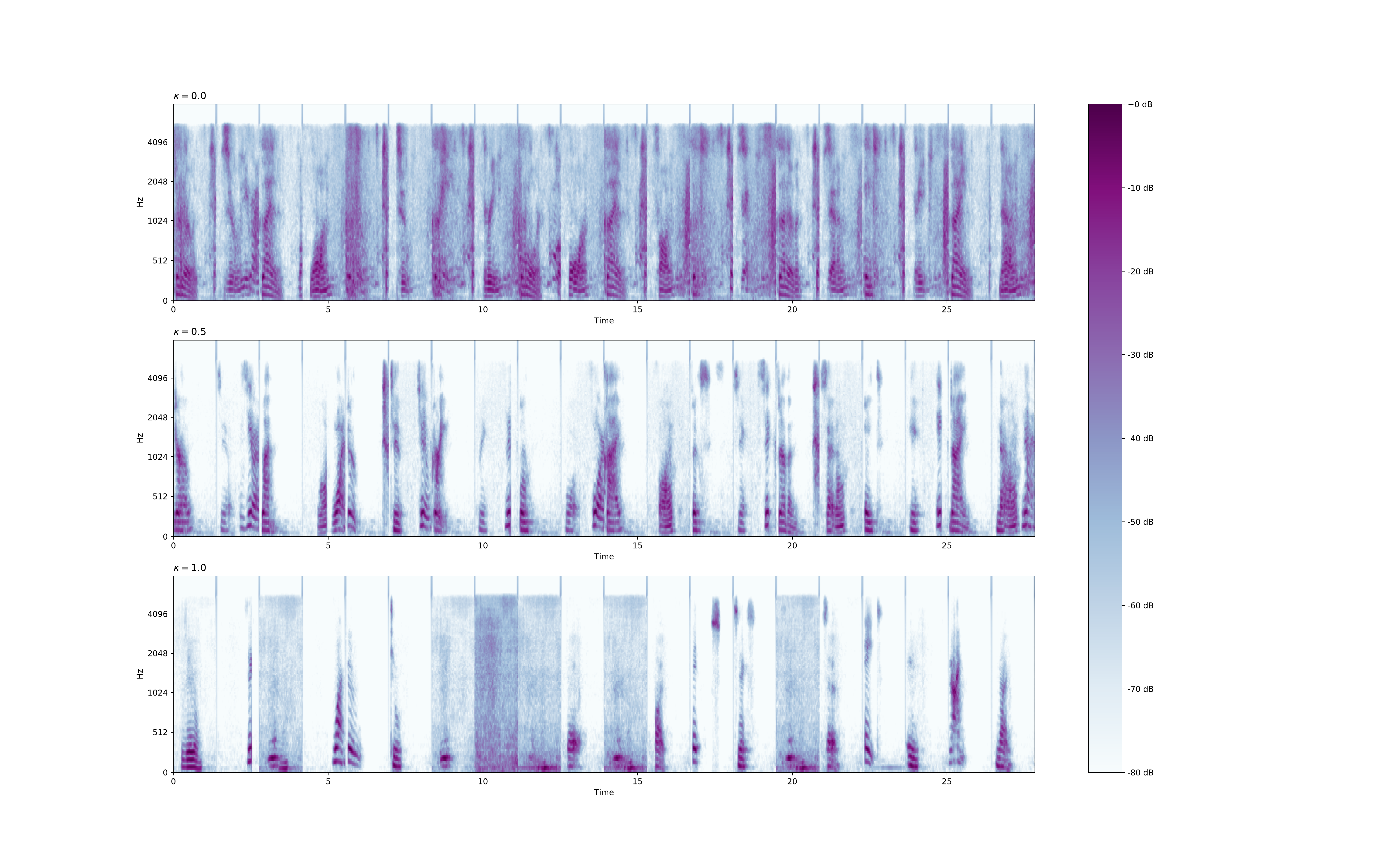}
    }\\
    \subfloat[][VAR + DDIM-rev ($\kappa=0.0$) (top) / DDIM-rev ($\kappa=0.5$) (middle) / DDPM-rev (bottom)]{
        \includegraphics[trim=150 40 270 60, clip, width=0.85\textwidth]{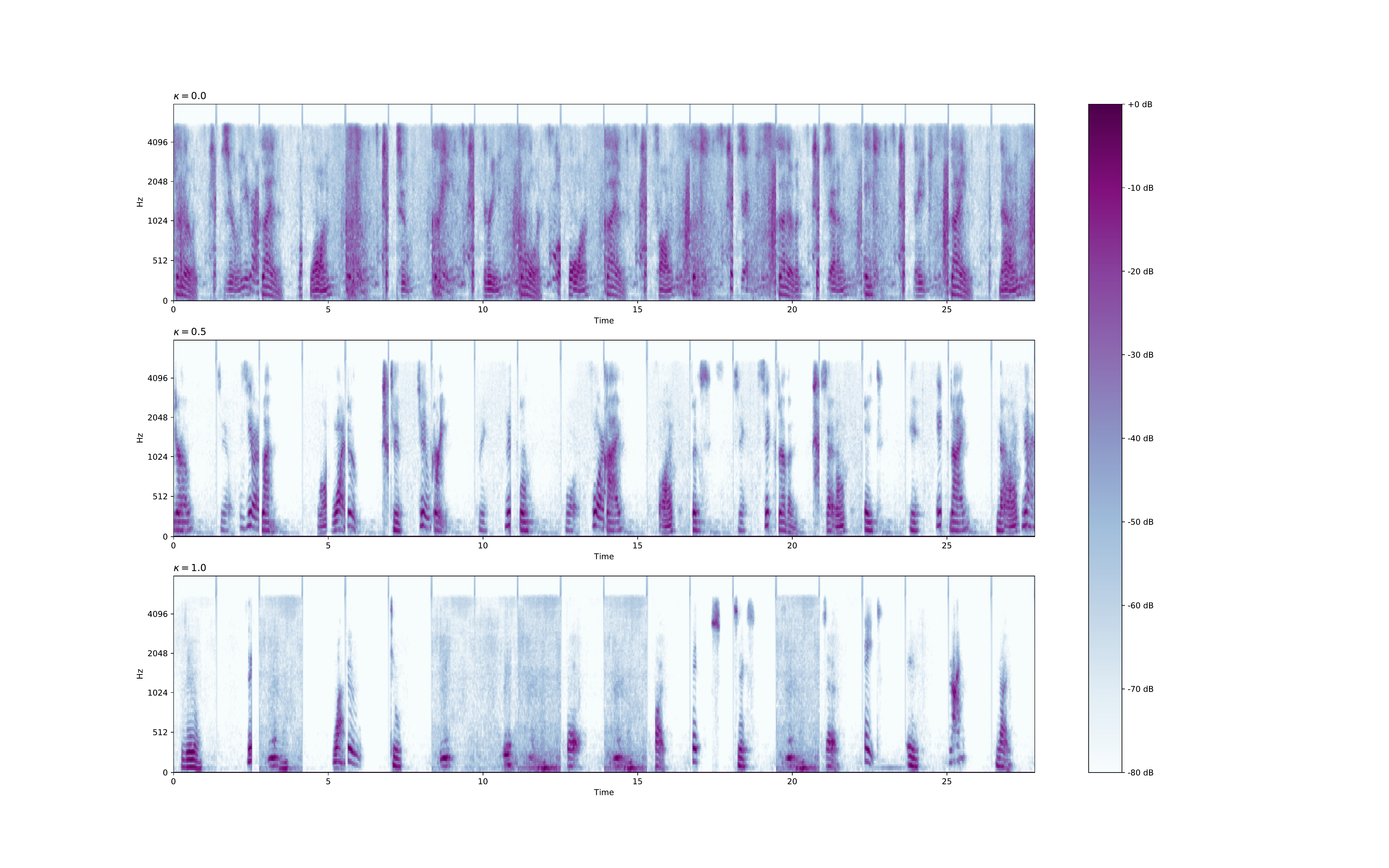}
    }
    \caption{Mel-spectrogram of 20 synthesized utterances ($S=50$). We use linear noise level schedules from steps in (a) and variances in (b). In each subplot, the top row shows results of DDIM-rev ($\kappa=0.0$), the middle row shows results of DDIM-rev ($\kappa=0.5$), and the bottom row shows results of DDPM-rev. DDIM-rev ($\kappa=0.5$) produces the clearest utterances in these approximate reverse processes.}
    \label{fig: sc09 cond mel}
\end{figure}

\newpage
\subsection{Neural Vocoding on LJSpeech}

\begin{figure}[!h]
    \centering
    \subfloat[][Ground truth (top) / STEP + DDIM-rev (middle) / STEP + DDPM-rev (bottom)]{
        \includegraphics[trim=150 80 270 60, clip, width=0.85\textwidth]{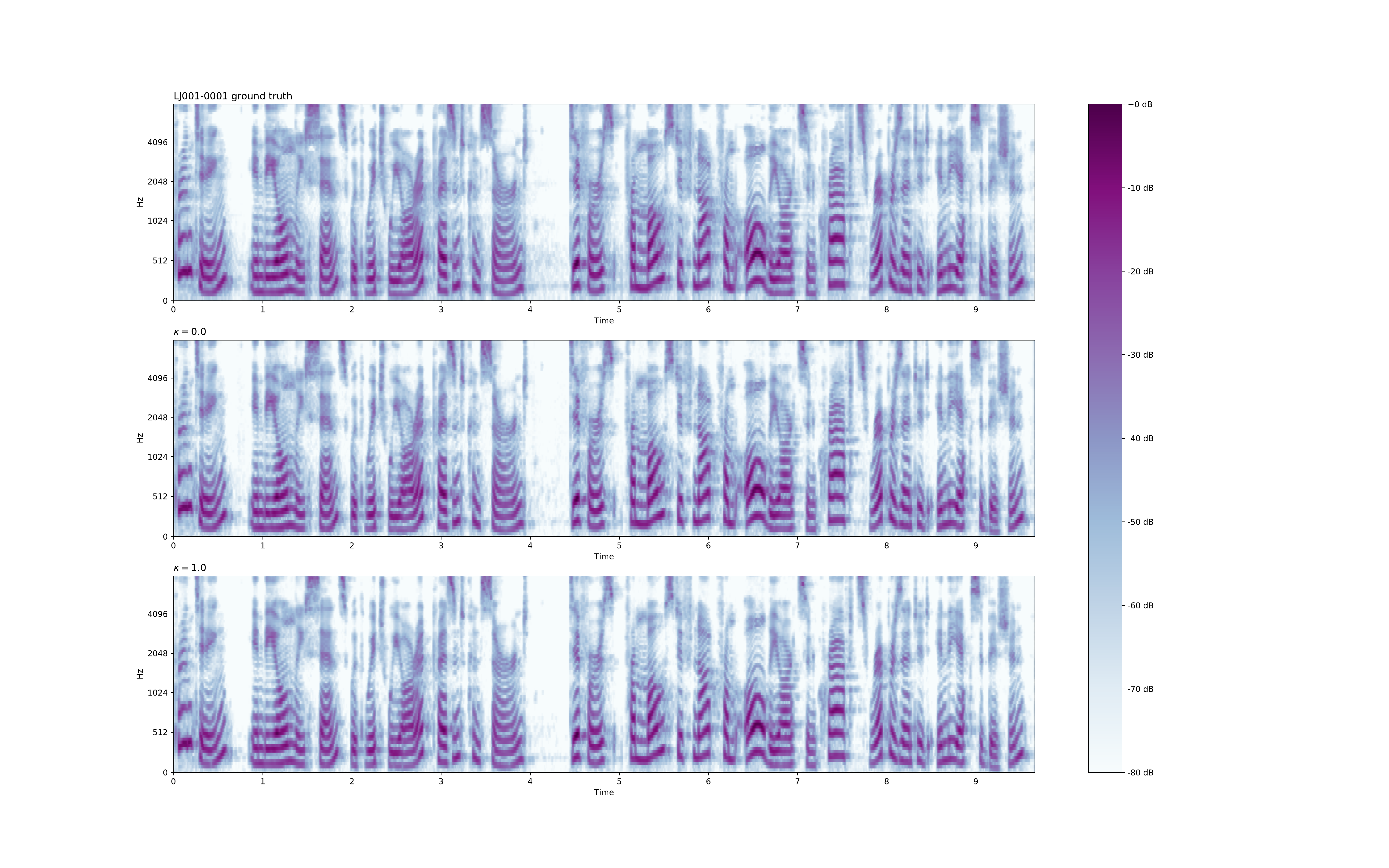}
    }\\
    \subfloat[][Ground truth (top) / VAR + DDIM-rev (middle) / VAR + DDPM-rev  (bottom)]{
        \includegraphics[trim=150 80 270 60, clip, width=0.85\textwidth]{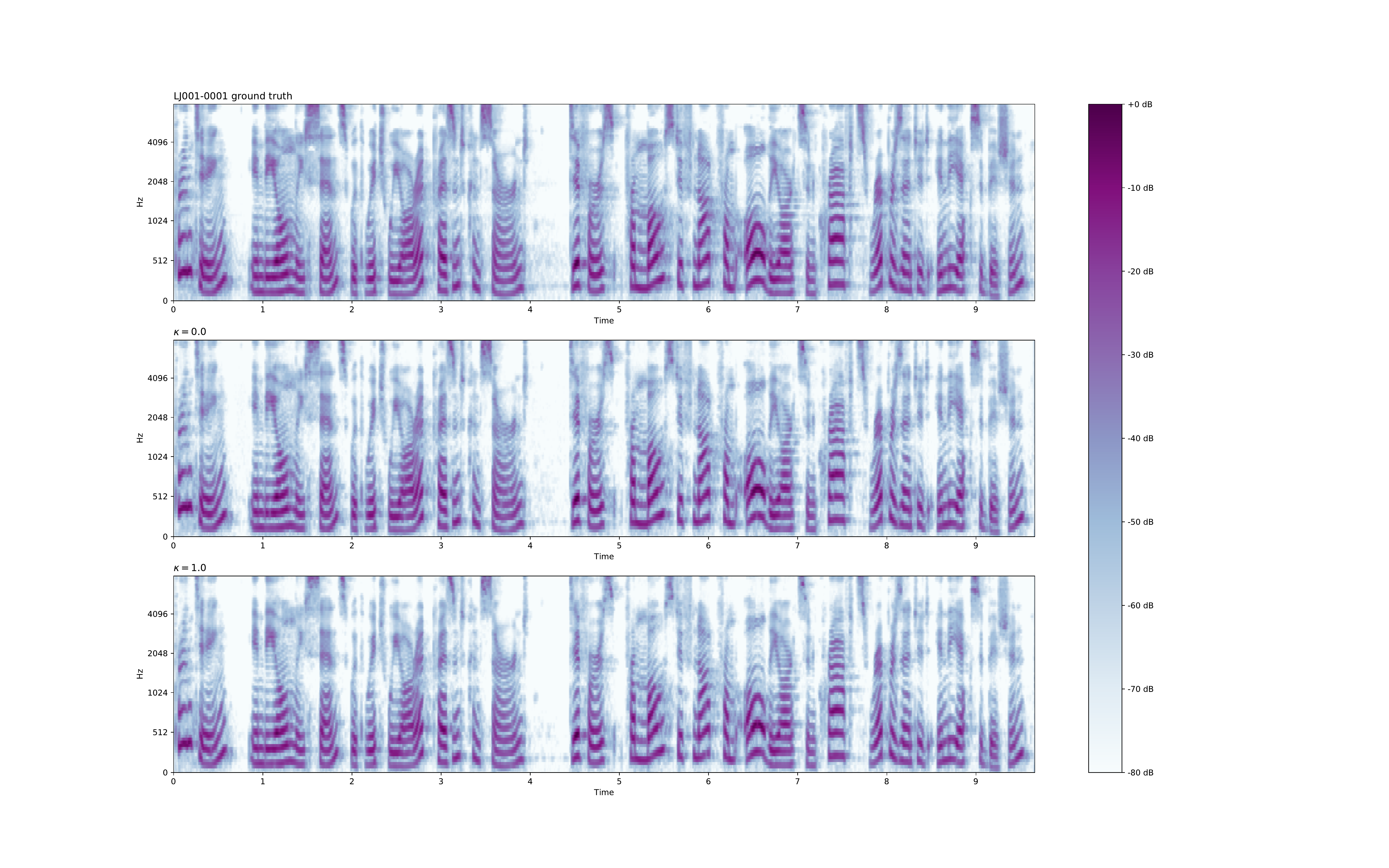}
    }
    \caption{Mel-spectrogram of ground truth and generated LJ001-0001 ($S=5$, channel$=128$). We use linear noise level schedules from steps in (a) and variances in (b). In each subplot, the top row shows ground truth, the middle row shows results of DDIM-rev ($\kappa=0.0$), and the bottom row shows results of DDPM-rev. Both DDPM-rev and DDIM-rev generate high quality speech.}
    \label{fig: vocoder mel}
\end{figure}

\end{document}

%% file: math_commands.tex
\usepackage{amsmath,amsfonts,bm}

\def\eqref#1{Eq.~(\ref{#1})}

\def\1{\bm{1}}

\DeclareMathAlphabet{\mathsfit}{\encodingdefault}{\sfdefault}{m}{sl}
\SetMathAlphabet{\mathsfit}{bold}{\encodingdefault}{\sfdefault}{bx}{n}

\def\gN{{\mathcal{N}}}
\def\gO{{\mathcal{O}}}

\def\gR{{\mathcal{R}}}

\def\gT{{\mathcal{T}}}

\newcommand{\qdata}{q_{\rm{data}}}

